\documentclass[letterpaper]{article} 
\usepackage{aaai23}  
\usepackage{times}  
\usepackage{helvet}  
\usepackage{courier}  
\usepackage[hyphens]{url}  
\usepackage{graphicx} 
\usepackage{natbib}  
\usepackage{caption} 
\usepackage{algorithm}
\usepackage{algorithmic}

\usepackage{threeparttable}
\usepackage{multirow}
\usepackage{multicol}
\usepackage{booktabs}
\usepackage{graphicx}
\usepackage{subfigure}
\usepackage{microtype}
\usepackage{url}
\usepackage{amssymb}
\usepackage{amsmath}
\usepackage{dsfont}
\usepackage[switch]{lineno}
\usepackage{xcolor}         

\usepackage{newfloat}
\usepackage{listings}
\DeclareCaptionStyle{ruled}{labelfont=normalfont,labelsep=colon,strut=off} 
\floatstyle{ruled}
\newfloat{listing}{tb}{lst}{}
\floatname{listing}{Listing}
\title{Controlling Class Layout for Deep Ordinal Classification\\ via Constrained Proxies Learning}
\author{
    Cong Wang, 
    Zhiwei Jiang\thanks{Corresponding author},
    Yafeng Yin,
    Zifeng Cheng,
    Shiping Ge,
    Qing Gu
}
\affiliations{
    State Key Laboratory for Novel Software Technology, Nanjing University, China \\

    cw@smail.nju.edu.cn, \{jzw, yafeng\}@nju.edu.cn, \{chengzf, shipingge\}@smail.nju.edu.cn, guq@nju.edu.cn
}

\usepackage{bibentry}

\begin{document}

\maketitle

\begin{abstract}
For deep ordinal classification, learning a well-structured feature space specific to ordinal classification is helpful to properly capture the ordinal nature among classes.
Intuitively, when Euclidean distance metric is used, an ideal ordinal layout in feature space would be that the sample clusters are arranged in class order along a straight line in space.
However, enforcing samples to conform to a specific layout in the feature space is a challenging problem.
To address this problem, in this paper, we propose a novel Constrained Proxies Learning (CPL) method, which can learn a proxy for each ordinal class and then adjusts the global layout of classes by constraining these proxies.
Specifically, we propose two kinds of strategies: hard layout constraint and soft layout constraint.
The hard layout constraint is realized by directly controlling the generation of proxies to force them to be placed in a strict linear layout or semicircular layout (i.e., two instantiations of strict ordinal layout).
The soft layout constraint is realized by constraining that the proxy layout should always produce unimodal proxy-to-proxies similarity distribution for each proxy (i.e., to be a relaxed ordinal layout).
Experiments show that the proposed CPL method outperforms previous deep ordinal classification methods under the same setting of feature extractor.
\end{abstract}

\section{Introduction}

Ordinal classification, also known as ordinal regression, aims to predict the label of samples on the ordinal scale.
It is a learning paradigm lying between multi-class classification and regression.
Compared with multi-class classification, the classes in ordinal classification are naturally ordered.
Compared with regression, the number of classes in ordinal classification is finite and the distance between adjacent classes is undefined.
Some examples include predicting the age group of a person (e.g., from \textit{Infants}, \textit{Children} to \textit{Aged}) and the star rating of a movie (e.g., from 1 star to 5 stars).

Traditional ordinal classification methods \cite{frank2001simple, chu2005gaussian, cardoso2007learning, lin2012reduction} mainly work on handcrafted features, which is labor-intensive and time-consuming.
Recently, with the great progress brought by deep neural networks, several deep ordinal classification methods have been proposed \cite{liu2018constrained, diaz2019soft, shaham2020deep, li2021learning} and show superior performance than traditional methods \cite{liu2018constrained}.
Such performance gain is mainly attributed to neural networks' strong capability of representation learning.

\begin{figure}[t]
    \centering
    \includegraphics[width=0.99\columnwidth]{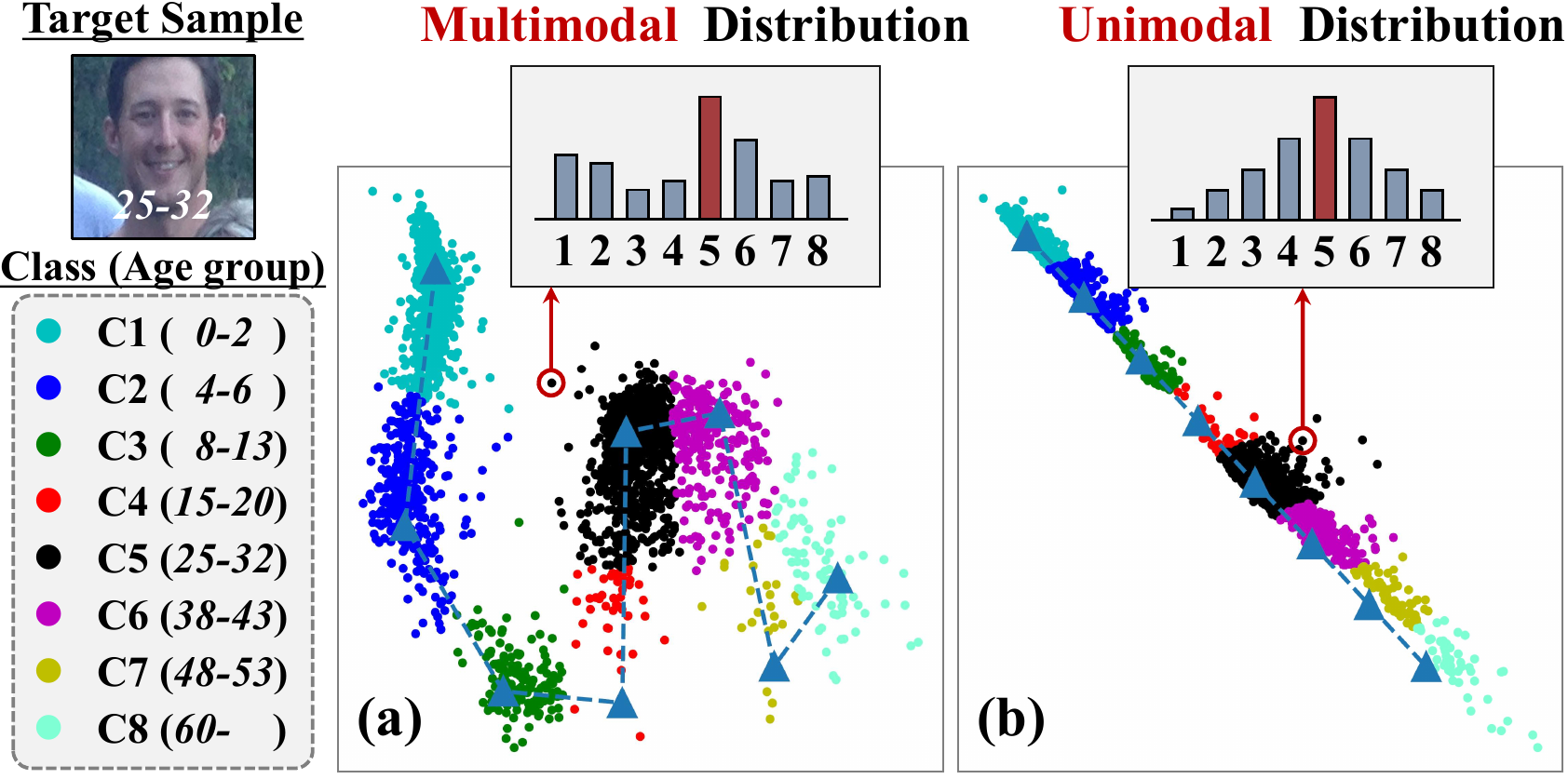}
    \caption{Illustration of (a) the unconstrained feature space and (b) our target ordinal constrained feature space.}
    \label{fig:paradigms}
\end{figure}

To benefit more from the representation learning, existing deep ordinal classification methods seek to learn the feature space specific to ordinal classification.
These methods fall into two fashions: classification and regression.
For the case of general multi-class classification, neither feature space nor output label distribution shows any ordinal property.
Therefore, researchers proposed to make implicit ordinal constraints on feature space by recoding the labels, such as transforming the $N$-class classification into $N-1$ binary classifications \cite{niu2016ordinal}, predicting the relative order of samples in triplets \cite{liu2018constrained}, and softening the one-hot label distribution to unimodal distribution \cite{diaz2019soft}.
For the case of regression, the samples are directly mapped into a one-dimensional space (i.e., a line of real numbers), which is ordered in nature.
But the samples are regressed into real numbers, which need to be discretized into classes by the learned boundaries, based on the unimodal transformation layer \cite{beckham2017unimodal} or Gaussian process layer \cite{liu2019probabilistic}.
The main difference between these two fashions of methods is that, the classification fashion constrains the feature space in a soft way by constraining the output label distribution, while the regression fashion constrains the feature space in a hard way by utilizing the ordinal nature of one-dimensional space.

Inspired by these work, we consider whether we can explicitly constrain the global layout of samples in the feature space to make it reflect the ordinal nature of classes.
Different from the regression fashion, which is only applicable in the one-dimensional space, we expect to do so in feature space of any dimension.
When a feature space is unconstrained, as shown in Figure~\ref{fig:paradigms}(a), the layout of samples can hardly guarantee the ordinal nature of classes.
With such layout, the samples of some faraway classes may be closely distributed in space (e.g., class 1 and class 5 in Figure~\ref{fig:paradigms}(a)), which may result in multimodal probability distributions for some samples (e.g., the target sample in Figure~\ref{fig:paradigms}(a)).
But if a feature space is ordinal constrained and the sample clusters are arranged in class order along a straight line in space, as shown in Figure~\ref{fig:paradigms}(b), samples can always get the unimodal probability distribution (when using Euclidean distance metric).
Previous studies have shown that the unimodal distribution is the ideal probability distribution for ordinal classification 
\cite{beckham2017unimodal}, which effectively reflects the ordinal nature of classes.
Thus, with such ordinal constrained layout, ordinal nature of classes can be guaranteed.

However, enforcing samples to conform to a specific layout in the feature space is a challenging problem.
To address this problem, we propose a novel Constrained Proxies Learning method (CPL), which can learn a proxy for each ordinal class and then adjusts the layout of classes by constraining these proxies.
Similar with previous two fashions, we explore two strategies of constraining the layout of proxies: hard layout constraint and soft layout constraint.
For the hard layout constraint, we directly control the generation of proxies to force them to be placed in an ordinal layout.
Considering that the ordinal layout is different under different metrics, we provide two instantiations of hard layout constraint: a linear layout specific to the Euclidean distance metric, and a semicircular layout specific to the cosine similarity metric.
For the soft layout constraint, we constrain the distribution shape of the similarities between each proxy and all proxies to be unimodal, so that the class layout can be a relaxed ordinal layout corresponding to a specific unimodal function.

Our main contributions are summarized as follows:
\begin{itemize}
    \item We propose a constrained proxies learning method to explicitly control the global layout of classes in high-dimensional feature space, making it more suitable for ordinal classification.
    \item We propose both the hard and soft layout constraints of proxies, and explore some example layouts for both of them (i.e., two strict ordinal layouts for hard constraint and two relaxed ordinal layouts for soft constraint).
    \item We conduct experiments\footnote{Code is available at https://github.com/tenvence/cpl.} on three public datasets and show that the proposed CPL achieves better performance than previous ordinal classification methods.
\end{itemize}

\section{Related Work}

\vspace{0.3em}
\noindent\textbf{Deep Ordinal Classification.}
Research on ordinal classification has last for about half a century \cite{gutierrez2015ordinal, mccullagh1980regression}.
Among these studies, deep ordinal classification has drawn a lot of attention in recently years, which mainly solve the task from two fashions: classification and regression.
The methods fall into classification fashion usually capture the ordinal information among classes by recoding the labels.
\citet{niu2016ordinal} transformed the $K$-class ordinal classification task to $K-1$ binary classification tasks.
Thus, the label can be recoded as a vector with $K-1$ dimensions, where the value of dimension $k$ is the answer of \textit{Is the target rank greater than $k$}.
\citet{liu2018constrained} trained their models on the triplets sampled from 3 adjacent classes by computing a pairwise hinge loss.
Thus, the target label can be recoded as a vector with $K$ dimensions, where the value of dimension $k$ is the answer of \textit{Is the target rank greater than $k-1$ and smaller than $k+1$}.
\citet{diaz2019soft} trained their model with general classification paradigm, but the target one-hot label vector is softened to the label distribution with unimodal shape.
\citet{shaham2020deep} used the modified proportional odds model to ensure that the output distribution is unimodal.
The methods fall into regression usually first map the samples into real numbers, then predict their class by learning the boundaries between classes.
\citet{beckham2017unimodal} transformed the output real number into an unimodal distribution, and predict the class by softmax operation.
\citet{liu2019probabilistic} designed a Gaussian Process layer to map the samples and learn the boundaries of classes.
\citet{li2021learning} proposed probabilistic ordinal embeddings (POEs) to exploit the ordinal nature of regression.
While these methods constrain the feature space implicitly, we provide a way of explicitly constraining the global layout of classes to be ordinal layout in feature space.

\begin{figure*}[t]
    \centering
    \includegraphics[width=2.05\columnwidth]{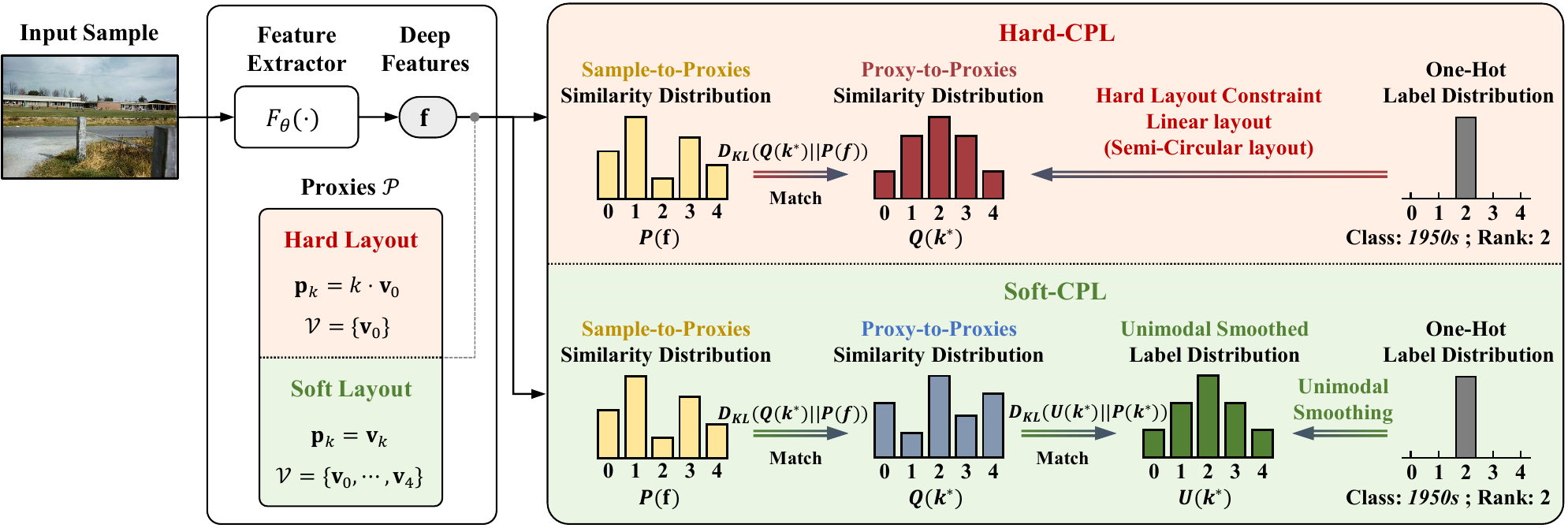}
    \caption{The framework of Constrained Proxies Learning (CPL) for deep ordinal classification.}
    \label{fig:framework}
\end{figure*}

\vspace{0.3em}
\noindent\textbf{Proxies Learning.}
Proxies learning is a novel paradigm of metric learning.
Previous metric learning methods \cite{bromley1993signature,  chopra2005learning,  hoffer2015deep, oh2017deep} mainly focus on constructing training pairs.
To get rid of the dependence on these sampling strategies and accelerate the convergence during training, Proxy-NCA \cite{movshovitz2017no} was proposed where each class is assigned with a learnable proxy, and the model can be trained in a way of single input.
Based on Proxy-NCA, SoftTriple loss \cite{qian2019softtriple} was proposed to assign multiple proxies to each category to reflect intra-class variance.
Manifold Proxy loss \cite{aziere2019ensemble} extended N-pair \cite{sohn2016improved} loss using proxies, and replaces Euclidean distance with a manifold-aware distance to improve the performance.
Recently, Proxy-Anchor \cite{kim2020proxy} was proposed to reformulate the triplet loss by viewing the proxy as anchor, and achieves good performance.
While existing proxies learning methods often focus on modeling the local relationship between samples and proxies, the research on explicitly constraining the global layout of proxies to a specific layout is still rare.

\section{Method}

For the ordinal classification task with $K$ classes, each sample $x$ belongs to a class $r_k\in\mathcal{R}$, where $\mathcal{R}=\{r_0, r_1, \cdots, r_{K-1}\}$ is the set of all classes and there is an ordinal relationship among these classes $r_0\prec r_1\prec\cdots\prec r_{K-1}$.
The objective is to correctly classify each sample $x$ into the class $r_k$ it belongs to, while reducing the errors on the ordinal scale as many as possible.
In this paper, we tackle the task from a proxies learning manner \cite{movshovitz2017no} and propose the Constrained Proxies Learning (CPL) method.

\subsection{Framework of Constrained Proxies Learning}

Our CPL is designed based on the proxies learning, which can learn a proxy for each class in feature space so as to make samples belonging to the same class can be closely clustered together around the corresponding proxy.
But unlike general proxies learning method, where the proxies are learned freely without constraint, our CPL aims to constrain the global layout of proxies in feature space to make it more suitable for ordinal classification.
Specifically, two strategies of layout constraint are considered: hard layout constraint (Hard-CPL) and soft layout constraint (Soft-CPL).
For Hard-CPL, proxies are constrained to be generated in a specific way so that they can be placed in a predefined ordinal layout.
For Soft-CPL, proxies are constrained to be placed in an ordinal layout corresponding to a specific unimodal distribution.

Before describing Hard-CPL and Soft-CPL in more details, we first introduce the framework of CPL.
As shown in Figure~\ref{fig:framework}, the CPL model contains two components: a feature extractor $F_\theta(\cdot)$ with parameters $\theta$, and a proxies learner $G_\mathcal{V}(\cdot)$ with $N$ learnable parameter vectors $\mathcal{V}=\{\mathbf{v}_0,\cdots,\mathbf{v}_{N-1}\}$.
Note that these parameters (i.e., $\theta$ and $\mathcal{V}$) can be trained together and $N$ can be different under different layout constraints.
Among these two components, the feature extractor can map a sample $x$ to the embedding feature $\mathbf{f}=F_\theta(x)\in\mathbb{R}^d$ with $d$ dimensions.
The proxies learner can be used to generate $K$ proxies $\mathcal{P}=\{\mathbf{p}_0,\mathbf{p}_1,\cdots,\mathbf{p}_{K-1}\}$, where the proxy $\mathbf{p}_k\in\mathbb{R}^d$ is corresponding to the ordinal class $r_k$.

To train the CPL model, the objective is to encourage $\mathbf{f}$ to be close to the target proxy $\mathbf{p}_{k^*}$ and to be far away from other proxies according to their relative ordinal distance with the target proxy in the feature space, where $k^*$ denote the ground-truth class rank of the input sample.

To reach this goal, we need to first specify a similarity function $\mathrm{sim}(\cdot,\cdot)$, so that the sample-to-proxies similarity distribution of class assignment $P(\mathbf{f})$ for the input sample $x$ can be calculated based on $\mathbf{f}$ and $\mathcal{P}$ by the softmax function:
\begin{equation}
    \label{eq:assignment-probability-distribution}
    P_k(\mathbf{f})=\frac{\exp(
        \mathrm{sim}(\mathbf{f},\mathbf{p}_k)
    )}{\sum_{k'=0}^{K-1}\exp(
        \mathrm{sim}(\mathbf{f},\mathbf{p}_{k'})
    )}
\end{equation}

Besides, we can also calculate the proxy-to-proxies similarity distribution $Q(k^*)$ by the softmax function:
\begin{equation}
    \label{eq:target-probability-distribution}
    Q_k(k^*)=\frac{\exp(\mathrm{sim}(\mathbf{p}_{k^*}, \mathbf{p}_k))}{\sum_{k'=0}^{K-1}\exp(\mathrm{sim}(\mathbf{p}_{k^*}, \mathbf{p}_{k'})}
\end{equation}

Then, by matching the distribution $P(\mathbf{f})$ and the distribution $Q(k^*)$, the basic loss function of CPL based on Kullback-Leibler (KL) divergence \cite{kullback1951information} can be defined for model training:
\begin{equation}
    \label{eq:loss-function-basic}
    \begin{aligned}
        \mathcal{L}_\mathrm{basic}({\mathbf{f},k^*})&
        =D_{\mathrm{KL}}[Q(k^*) \Vert P(\mathbf{f})]\\
        &=-\frac{1}{K}\sum_{k=0}^{K-1} Q_k(k^*)\log\frac{P_k(\mathbf{f})}{Q_k(k^*)}    
    \end{aligned}
\end{equation}

Once the CPL model is well-trained, the predicted class rank $\hat{k}$ of the input sample $x$ can be inferred by finding the proxy most similar to its embedding feature:
\begin{equation}
    \label{eq:rank-prediction}
    \hat{k}=\arg\max_k P_k(F_\theta(x))=\arg\max_k \mathrm{sim}(F_\theta(x), \mathbf{p}_k)
\end{equation}

In addition, for the similarity function $\mathrm{sim}(\cdot,\cdot)$, we consider two examples based on Euclidean distance and cosine similarity, respectively.
For Euclidean distance, based on Student's t-distribution \cite{van2008visualizing}, the similarity function can be formulated as:
\begin{equation}
    \label{eq:sim-eu}
    \mathrm{sim_E}(\mathbf{f},\mathbf{p}_k)=-\log(1+\Vert \mathbf{f}-\mathbf{p}_k \Vert^2)
\end{equation}
For cosine similarity, the similarity function is formulated as:
\begin{equation}
    \label{eq:sim-cosine}
    \mathrm{sim_C}(\mathbf{f},\mathbf{p}_k)=s\cdot\cos(\mathbf{f},\mathbf{p}_k)
    =s\cdot\frac{\mathbf{f}^\mathrm{T}\mathbf{p}_k}{\Vert \mathbf{f} \Vert \Vert \mathbf{p}_k \Vert}
\end{equation}
where $s$ is a hyperparameter to scale the range of cosine similarity and $\Vert\mathbf{p}_k\Vert=1$.
The scale parameter $s$ is limited as $s>1$, which is used in many cosine-based softmax loss \cite{deng2019arcface, huang2020curricularface, wang2018cosface, zhang2019adacos}.
Note that other similarity function can also be used in the CPL framework.

\subsection{Hard Layout Constrained Proxies Learning}

\begin{figure}[t]
    \centering\includegraphics[width=0.99\columnwidth]{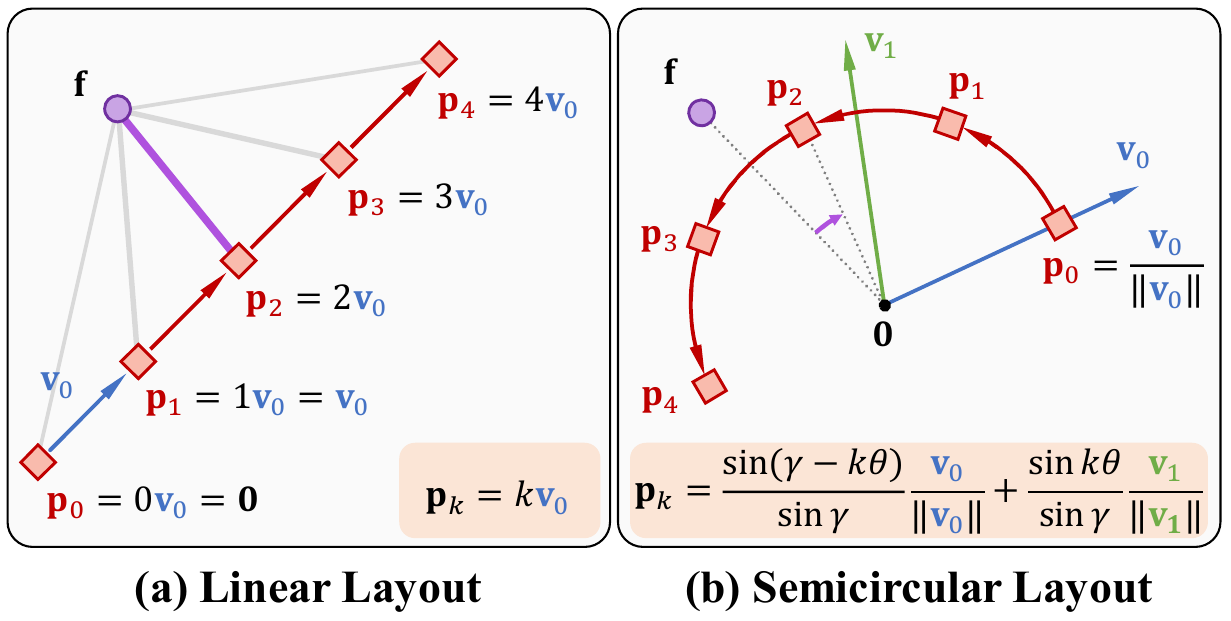}
    \caption{Two example schemes of Hard-CPL.}
    \label{fig:hard-cpl}
\end{figure}

For Hard-CPL, the layout of proxies is constrained in a hard way by directly constraining the generation of proxies.
As shown in Figure~\ref{fig:paradigms}, compared with unconstrained proxies learning, Hard-CPL constrains the proxies to be placed in a straight line and thus can achieve a strict ordinal layout.
Considering that such linear layout guarantees the ordinal nature of classes only if Euclidean distance is used as metric, we also provide another example scheme of ordinal layout specific to the cosine similarity, called semicircle layout.

\vspace{0.3em}
\noindent\textbf{Linear Layout for Euclidean Distance (H-L).}
Intuitively, a straight line in a space is determined by two points (vectors).
Thus, the linear layout constraint can be $\mathbf{p}_k=G_\mathcal{V}(k)=\mathbf{v}_0+k\cdot\mathbf{v}_1$ which means $N=2$ and $\mathcal{V}=\{\mathbf{v}_0,\mathbf{v}_1\}$.
But when computing the Euclidean distance $\Vert\mathbf{f}-\mathbf{p}_k\Vert=\Vert(\mathbf{f}-\mathbf{v}_0)-k\cdot\mathbf{v}_1\Vert$, $\mathbf{v}_0$ can be considered as a bias for $\mathbf{f}$.
Then, for simplification, the proxies learner can be formulated as:
\begin{equation}
    \label{eq:proxies-learner-of-linear-layout-cpl}
    \mathbf{p}_k=G_\mathcal{V}(k)=k\cdot\mathbf{v}_0
\end{equation}
where $N=1$ and $\mathcal{V}=\{\mathbf{v}_0\}$.
As illustrated in Figure~\ref{fig:hard-cpl}(a), $\mathbf{p}_0$ is located in the origin point.
All proxies are evenly distributed in the direction of $\mathbf{v}_0$ with the margin of $\Vert\mathbf{v}_0\Vert$.

\vspace{0.3em}
\noindent\textbf{Semicircular Layout for Cosine Similarity (H-S).}
Intuitively, a semicircular arc need to be determined by a plane of the feature space.
Thus we use two vectors to determine a semicircular arc, which means $N=2$ and $\mathcal{V}=\{\mathbf{v}_0, \mathbf{v}_1\}$.
Specifically, $\mathbf{v}_0$ is used to determine the direction of the first proxy $\mathbf{p}_0$, and $\mathbf{v}_1$ is used to determine the expanding direction of the arc, as shown in Figure~\ref{fig:hard-cpl}(b).
Since the angle between $\mathbf{p}_0$ and $\mathbf{p}_{K-1}$ is $\pi$ (i.e., a semicircular), the angle between adjacent proxies can be set to $\pi/(K-1)$ to equally divide the semicircle.
Then the proxies learner is formulated as:
\begin{equation}
    \label{eq:proxies-learner-of-semicircular-layout-cpl}
    \mathbf{p}_k=G_\mathcal{V}(k)=
    \frac{\sin(\gamma-k\beta)}{\sin\gamma}\cdot\frac{\mathbf{v}_0}{\Vert\mathbf{v}_0\Vert}
    +
    \frac{\sin k\beta}{\sin\gamma}\cdot\frac{\mathbf{v}_1}{\Vert\mathbf{v}_1\Vert}
\end{equation}
where $\Vert\mathbf{p}_k\Vert=1$, $\beta=\pi/(K-1)$, and $\gamma$ is the angle between $\mathbf{v}_0$ and $\mathbf{v}_1$, $\gamma=\arccos(
    \mathbf{v}_0^\mathrm{T}\mathbf{v}_1/
    \Vert\mathbf{v}_0\Vert\Vert\mathbf{v}_1\Vert
)$.

\subsection{Soft Layout Constrained Proxies Learning}

For Soft-CPL, we relax the hard layout constraint, allowing proxies not to be placed in strict linear/semicircular layout.
Therefore, we allow the proxies to be learned freely and only constrain that the proxy layout should always produce unimodal proxy-to-proxies similarity distribution for each proxy.

To this end, each proxy has its own learnable parameter vector, and the proxies learner can be formulated as:
\begin{equation}
    \label{eq:proxies-learner-of-general-proxies-learning}
    \mathbf{p}_k=G_\mathcal{V}(k)=\mathbf{v}_k
\end{equation}
where $\mathcal{V}=\{\mathbf{v}_0,\cdots,\mathbf{v}_{K-1}\}$.

To constrain the proxy-to-proxies similarity distribution $Q(k^*)$ to be unimodal, we can first define a unimodal smoothed label distribution $U(k^*)$ by a unimodal smoothing function $E(\cdot,\cdot)$ and the softmax function:
\begin{equation}
    \label{eq:unimodal-probability-distribution}
    U_k(k^*)=\frac{\exp(E(k;k^*))}{\sum_{k'=0}^{K-1}\exp(E(k';k^*))}
\end{equation}
Then, by matching the distributions $Q(k^*)$ and $U(k^*)$, we can define an extra unimodal loss function for Soft-CPL:
\begin{equation}
    \label{eq:loss-function-unimodal}
    \begin{aligned}
        \mathcal{L}_\mathrm{unimodal}({k^*})&
        =D_{\mathrm{KL}}[U(k^*) \Vert Q(k^*)]   
    \end{aligned}
\end{equation}
 While Hard-CPL only use the basic loss for model training:
\begin{equation}
    \label{eq:hard-loss}
    \mathcal{L}_\mathrm{H}=\mathcal{L}_\mathrm{basic}
 \end{equation}
 Soft-CPL use both the basic loss and unimodal loss:
\begin{equation}
    \label{eq:soft-loss}
    \mathcal{L}_\mathrm{S}=\mathcal{L}_\mathrm{basic}+\alpha\mathcal{L}_\mathrm{unimodal}
  \end{equation}
where $\alpha$ is tradeoff parameter.
 
For the unimodal smoothing function $E(\cdot,\cdot)$, we consider two classic unimodal distributions \cite{beckham2017unimodal} as examples: Poisson distribution and Binomial distribution.
Note that other unimodal distributions are also applicable.

\vspace{0.3em}
\noindent\textbf{Poisson Distribution (S-P).}
The log probability mass function (LPMF) of Poisson distribution is defined as:
\begin{equation}
    \label{eq:lpmf-poisson}
    \mathrm{LPMF}(k;\lambda)=k\log\lambda-\lambda-\log k!
\end{equation}
where $k\in\mathbb{N}$, $\lambda\in\mathbb{R}^+$, and the maximum value of LPMF is taken when the condition $k<\lambda<k+1$ is met.
Then the ordinal smoothing function $E$ can be formulated as:
\begin{equation}
    \label{eq:tef-poisson}
    E(k;k^*)=\frac{1}{\tau_p}\cdot\mathrm{LPMF}\left(k;k^*+\frac{1}{2}\right)
\end{equation}
where $\tau_p\in(0,+\infty)$ is a hyperparameter to control the shape of the distribution.
When $\tau_p\rightarrow +\infty$ or $\tau_p\rightarrow 0$, the distribution will tend to be a uniform distribution or a one-hot distribution respectively.

\begin{table*}[t] \footnotesize \setlength{\tabcolsep}{9.4pt}
    \centering
    \begin{tabular}{lllcccc}
        \toprule
        \multicolumn{3}{l}{\multirow{2}{*}{\textbf{Methods}}} & \multicolumn{2}{c}{\textbf{Historical Color}} & \multicolumn{2}{c}{\textbf{Adience Face}} \\ 
        \cmidrule(lr){4-5} \cmidrule(lr){6-7}
        & & & \textbf{Accuracy (\%) $\uparrow$} & \textbf{MAE} $\downarrow$ & \textbf{Accuracy (\%)} $\uparrow$ & \textbf{MAE} $\downarrow$ \\
        \midrule
        \multicolumn{3}{l}{\textbf{Classification} \cite{liu2018constrained}} &
        48.94 $\pm$ 2.54 & 0.89 $\pm$ 0.06 & 54.0 $\pm$ 6.3 & 0.61 $\pm$ 0.08 \\ 
        \multicolumn{3}{l}{\textbf{Regression} \cite{niu2016ordinal}} & 
        42.24 $\pm$ 2.91 & 0.79 $\pm$ 0.03 & 56.3 $\pm$ 4.9 & 0.56 $\pm$ 0.07 \\
        \multicolumn{3}{l}{\textbf{Ranking} \cite{li2021learning}} & 
        44.67 $\pm$ 4.24 & 0.81 $\pm$ 0.06 & 56.7 $\pm$ 6.0 & 0.54 $\pm$ 0.08 \\
        \midrule
        \multicolumn{3}{l}{\textbf{CNNPOR} \cite{liu2018constrained}} & 
        50.12 $\pm$ 2.65 & 0.82 $\pm$ 0.05 & 57.4 $\pm$ 5.8 & 0.55 $\pm$ 0.08 \\
        \multicolumn{3}{l}{\textbf{GP-DNNOR} \cite{liu2019probabilistic}} & 
        46.60 $\pm$ 2.98 & 0.76 $\pm$ 0.05 & 57.4 $\pm$ 5.5 & 0.54 $\pm$ 0.07 \\
        \multicolumn{3}{l}{\textbf{SORD} \cite{diaz2019soft}} & 
        -- & -- & 59.6 $\pm$ 3.6 & 0.49 $\pm$ 0.05 \\
        \multicolumn{3}{l}{\textbf{POEs} \cite{li2021learning} } &
        54.68 $\pm$ 3.21 & 0.66 $\pm$ 0.05 & 60.5 $\pm$ 4.4 & 0.47 $\pm$ 0.06 \\ \midrule \midrule
        
        \multicolumn{2}{l}{\multirow{2}{*}{\textbf{UPL}}} & \textbf{Euclidean Distance} & 
        52.20 $\pm$ 3.84 & 0.71 $\pm$ 0.07 & 58.1 $\pm$ 3.2 & 0.48 $\pm$ 0.05 \\
        
        & & \textbf{Cosine Similarity} & 
        51.32 $\pm$ 2.99 & 0.74 $\pm$ 0.05 & 56.8 $\pm$ 4.5 & 0.51 $\pm$ 0.07 \\ \midrule
        
        \multirow{6}{*}{\textbf{CPL}} & 
         \textbf{Hard-Linear} & \textbf{Euclidean Distance} & 
        55.71 $\pm$ 3.20 & \textbf{0.63 $\pm$ 0.06} & 61.6 $\pm$ 2.6 & \textbf{0.43 $\pm$ 0.04} \\ 
        
        & \textbf{Hard-Semicircular} & \textbf{Cosine Similarity} & 
        55.41 $\pm$ 3.21 & \underline{0.64 $\pm$ 0.06} & 61.8 $\pm$ 3.1 & \textbf{0.43 $\pm$ 0.04} \\ \cmidrule{2-7}
         
        & \multirow{2}{*}{\textbf{Soft-Poisson}} & \textbf{Euclidean Distance} & 
        57.28 $\pm$ 3.41 & 0.65 $\pm$ 0.07 & 61.3 $\pm$ 3.7 & 0.45 $\pm$ 0.05 \\
        
        & & \textbf{Cosine Similarity} &
        56.99 $\pm$ 2.44 & 0.65 $\pm$ 0.05 & 61.1 $\pm$ 4.0 & 0.46 $\pm$ 0.05 \\ \cmidrule{2-7}
        
        & \multirow{2}{*}{\textbf{Soft-Binomial}} & \textbf{Euclidean Distance} & 
        \textbf{57.96 $\pm$ 3.14} & 0.66 $\pm$ 0.08 & \textbf{62.1 $\pm$ 3.6} & \underline{0.44 $\pm$ 0.04} \\
        
        & & \textbf{Cosine Similarity} & 
        \underline{57.66 $\pm$ 3.11} & 0.65 $\pm$ 0.06 & \underline{61.9 $\pm$ 4.5} & \underline{0.44 $\pm$ 0.05} \\
        \bottomrule
    \end{tabular}
    \caption{The performance (accuracy and MAE) of all comparison methods on Historical Color dataset and Adience Face dataset. 
    The feature extractors are all VGG-16. 
    The best measures are in bold, and the second best measures are underlined.}
    \label{tbl:comparison-historical-color-adience-face}
\end{table*}

\begin{table*}[t] \footnotesize \setlength{\tabcolsep}{2.45pt}
    \centering
    \begin{tabular}{lllcccccccccc}
        \toprule
        \multicolumn{3}{l}{\multirow{2}{*}{\textbf{Methods}}} & \multicolumn{5}{c}{\textbf{Accuracy (\%)} $\uparrow$} & \multicolumn{5}{c}{\textbf{MAE} $\downarrow$}  \\ 
        \cmidrule(lr){4-8} \cmidrule(lr){9-13}
        &&& \textbf{Nature} & \textbf{Animals} & \textbf{Urban} & \textbf{People} & \textbf{Overall} & \textbf{Nature} & \textbf{Animals} & \textbf{Urban} & \textbf{People} & \textbf{Overall} \\ 
        \midrule
        \multicolumn{3}{l}{\textbf{Classification} \cite{liu2018constrained}} & 
        70.97 & 68.02 & 68.19 & 71.63 & 69.45 & 0.305 & 0.342 & 0.374 & 0.412 & 0.376 \\
        \multicolumn{3}{l}{\textbf{Regression} \cite{niu2016ordinal}} &
        71.52 & 70.72 & 71.22 & 69.72 & 70.80 & 0.378 & 0.397 & 0.387 & 0.400 & 0.390 \\ 
        \multicolumn{3}{l}{\textbf{Ranking} \cite{li2021learning}} & 
        69.81 & 69.10 & 66.49 & 66.49 & 68.96 & 0.313 & 0.331 & 0.349 & 0.312 & 0.326 \\ 
        \midrule
        \multicolumn{3}{l}{\textbf{CNNPOR} \cite{liu2018constrained}} &
        71.86 & 69.32 & 69.09 & 69.94 & 70.05 & 0.294 & 0.322 & 0.325 & 0.321 & 0.316 \\ 
        \multicolumn{3}{l}{\textbf{SORD} \cite{diaz2019soft}} &
        73.59 & 70.29 & \underline{73.25} & 70.59 & 72.03 & 0.271 & 0.308 & \textbf{0.276} & 0.309 & 0.290 \\
        \multicolumn{3}{l}{\textbf{POEs} \cite{li2021learning}} &
        73.62 & 71.14 & 72.78 & 72.22 & 72.44 & 0.273 & 0.299 & \underline{0.281} & 0.293 & 0.287 \\ 
        \midrule\midrule
        
        \multicolumn{2}{l}{\multirow{2}{*}{\textbf{UPL}}} & \textbf{Euclidean Distance} & 
        71.82 &	68.21 &	69.24 &	68.98 &	69.56 &	0.283 &	0.343 &	0.313 &	0.341 &	0.320 \\
        & & \textbf{Cosine Similarity} & 
        72.88 &	68.68 &	69.88 &	69.81 &	70.31 &	0.284 &	0.325 &	0.311 &	0.352 &	0.318 \\ \midrule
        
        \multirow{6}{*}{\textbf{CPL}} & 
        \textbf{Hard-Linear} & \textbf{Euclidean Distance} & 
        74.43 & 72.11 & 72.99 & 72.53 & 73.02 & \textbf{0.260} & \textbf{0.289} & 0.283 & \underline{0.287} & \textbf{0.280} \\
        
        & \textbf{Hard-Semicircular} & \textbf{Cosine Similarity} & 
        74.35 & 71.50 & 72.91 & 72.33 & 72.77 & \underline{0.262} & \underline{0.297} & 0.288 & 0.290 & \underline{0.284} \\  \cmidrule{2-13}
        
        & \multirow{2}{*}{\textbf{Soft-Poisson}} & \textbf{Euclidean Distance} &
        74.46 & 71.73 & 72.94 & 72.45 & 72.90 & 0.267 & 0.302 & \underline{0.281} & 0.297 & 0.287 \\ 
        
        & & \textbf{Cosine Similarity} &
        74.53 & 71.39 & 72.97 & 72.38 & 72.82 & 0.270 & 0.299 & 0.287 & \textbf{0.286} & 0.286 \\ \cmidrule{2-13}
        
        & \multirow{2}{*}{\textbf{Soft-Binomial}} & \textbf{Euclidean Distance} & 
        \textbf{74.97} & \textbf{72.61} & \textbf{73.28} & \underline{72.61} & \textbf{73.37} & \underline{0.262} & \underline{0.297} & 0.285 & 0.299 & 0.286 \\
        
        & & \textbf{Cosine Similarity} & 
        \underline{74.62} & \underline{72.28} & 73.20 & \textbf{72.74} & \underline{73.21} & 0.265 & 0.301 & 0.286 & 0.294 & 0.287 \\ 
        \bottomrule
    \end{tabular}
    \caption{The performance (accuracy and MAE) of all comparison methods on Image Aesthetics dataset. 
    The feature extractors are all VGG-16.
    The best measures are in bold, and the second best measures are underlined.}
    \label{tbl:comparison-image-aesthetics}
\end{table*}

\vspace{0.3em}
\noindent\textbf{Binomial Distribution (S-B).}
The LPMF of Binomial distribution is defined as:
\begin{equation}
    \label{eq:plmf-binomial}
    \begin{aligned}
        \mathrm{LPMF}(k;K-1,p)=&\log\binom{K-1}{k}+ k\log p\\
        & + (K-1-k)\log(1-p)
    \end{aligned}
\end{equation}
where $0\leq k\leq K-1$, $p\in[0,1]$, and the maximum value of LPMF is taken when the condition $k<Kp<k+1$ is met.
Then the ordinal smoothing function $E(\cdot,\cdot)$ can be formulated as:
\begin{equation}
    \label{eq:tef-binomial}
    E(k,k^*) = \frac{1}{\tau_b}\cdot\mathrm{LPMF}\left(k;K-1,\frac{2k+1}{2K}\right)
\end{equation}
where $\tau_b\in(0,+\infty)$ is a hyperparameter to control the shape.
Same as $\tau_p$ in Equation~(\ref{eq:tef-poisson}), $\tau_b$ is used to control the shape of the distribution between the uniform distribution and the one-hot distribution.

\section{Experiments}

\subsection{Datasets and Evaluation Metrics}

We employ three public datasets for evaluation, which are:

\begin{itemize}
    \item \textbf{Historical Color} \cite{palermo2012dating} is a small and balanced ordinal classification dataset which contains images captured on five decades, from \textit{1930s} to \textit{1970s}, each of which has 265 images.
    In each class, 210 images are randomly selected for training.
    For the rest 55 images, randomly selected 5 images are used for validation and 50 images are used for testing.
    The experiments are repeated 10 times with different partitions.
    \item \textbf{Adience Face} \cite{levi2015age} contains 26,580 face photos from 2,284 subjects.
    The dataset is divided into 8 age groups, which are \textit{0-2}, \textit{4-6}, \textit{8-13}, \textit{15-20}, \textit{25-32}, \textit{38-43}, \textit{48-53}, and \textit{elder than 60 years old}, respectively. The five-fold partition follows the official repository\footnote{\url{https://github.com/GilLevi/AgeGenderDeepLearning}}.
    \item \textbf{Image Aesthetics} \cite{schifanella2015image} provides 15,687 Flickr image URLs, while 13,774 images are available online. 
    The dataset contains four categories of images, namely nature, animals, people, and urban.
    The quality of each image is scored by at least five graders on the five scales, i.e., \textit{unacceptable}, \textit{flawed}, \textit{ordinary}, \textit{professional}, and \textit{exceptional}.
    In the experiments, we randomly select 75\% images as the training set, 5\% images as the validation set, and 20\% images as the test set.
    The experiments are repeated 5 times with different partitions.
\end{itemize}

The evaluation metrics are accuracy and Mean Absolute Error (MAE), which are widely-used in previous work.

\begin{figure*}[t]
    \centering
    \subfigure[$s$ for H-S]{
        \label{fig:param-scale}
        \includegraphics[height=0.41\columnwidth]{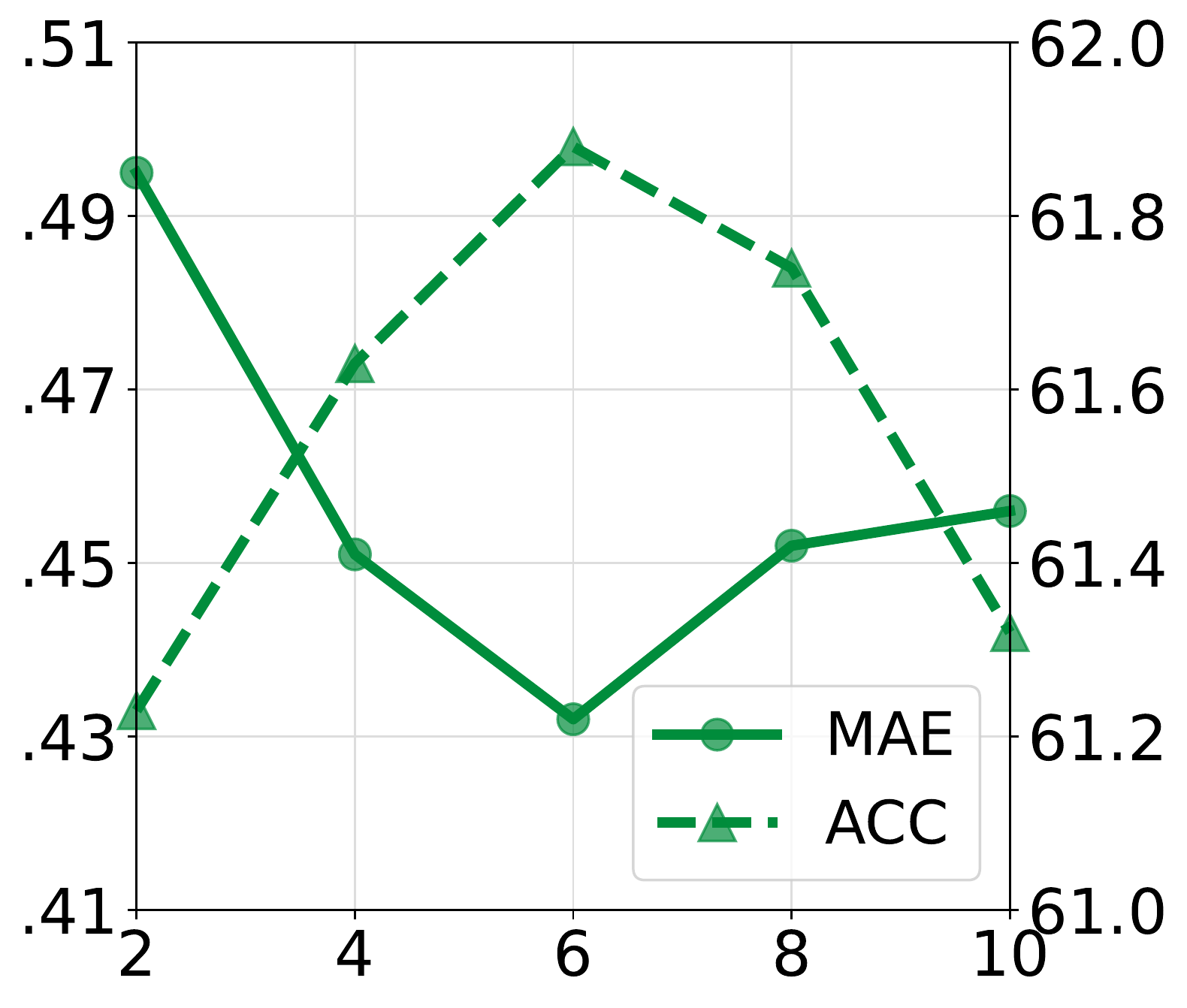}
    }
    \hfill
    \subfigure[$\tau_p$ and $\tau_b$ for S-P and S-B]{
        \label{fig:param-tau}
        \includegraphics[height=0.41\columnwidth]{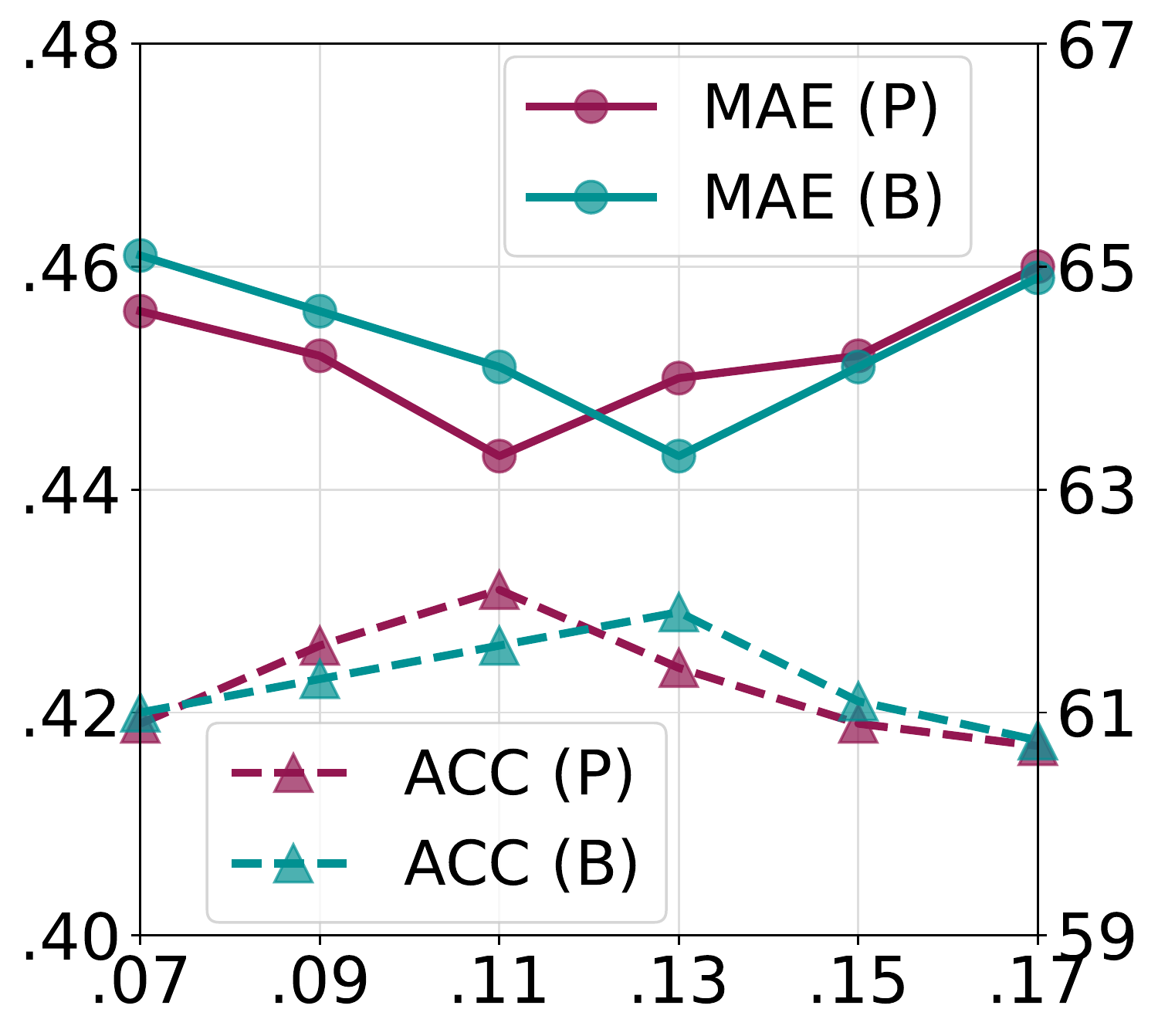}
    } 
    \hfill
    \subfigure[$\alpha$ for S-P and S-B]{
        \label{fig:param-lambda}
        \includegraphics[height=0.41\columnwidth]{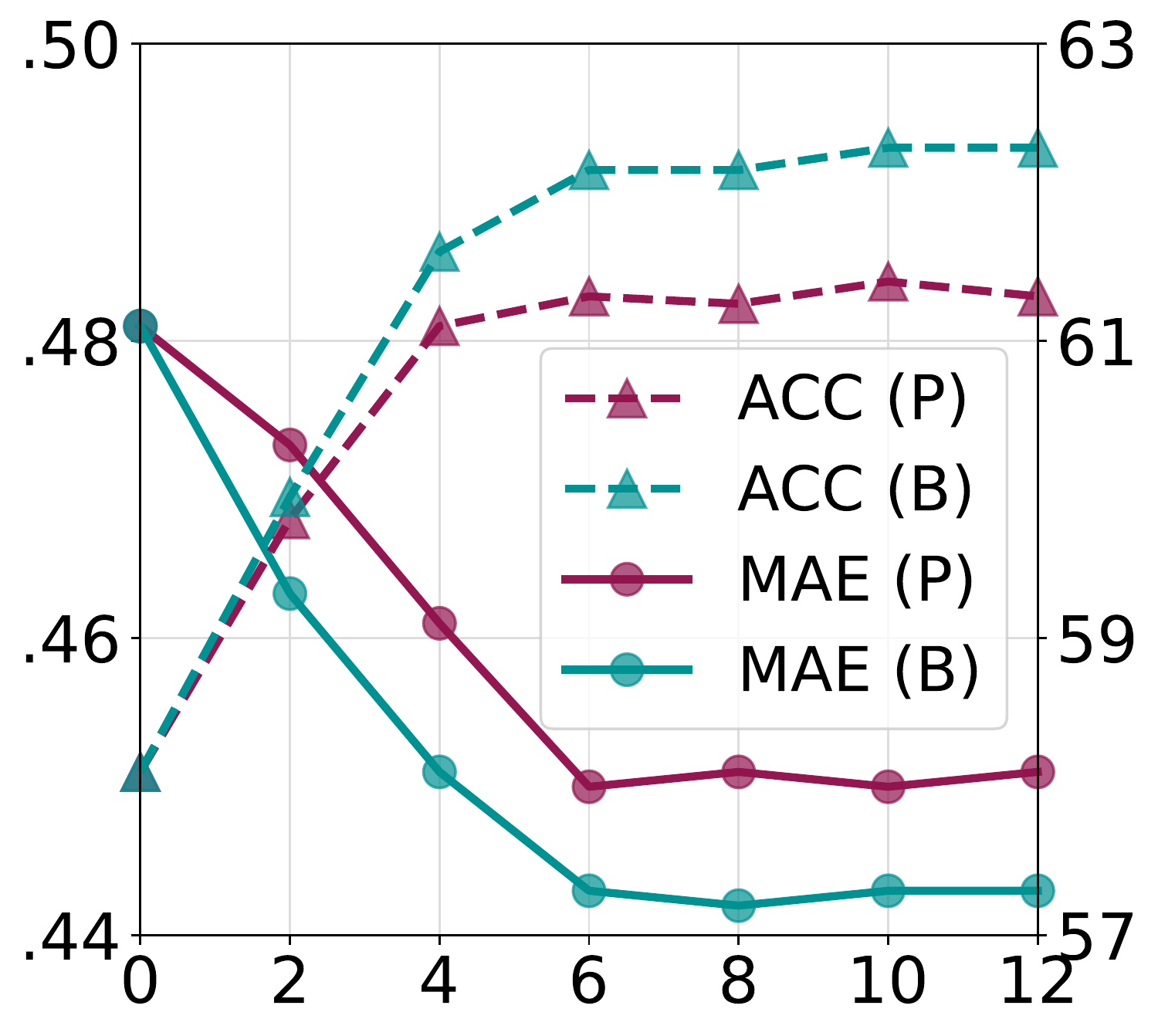}
    }
    \hfill
    \subfigure[Feature dimension]{
        \label{fig:feature-dim}
        \includegraphics[height=0.41\columnwidth]{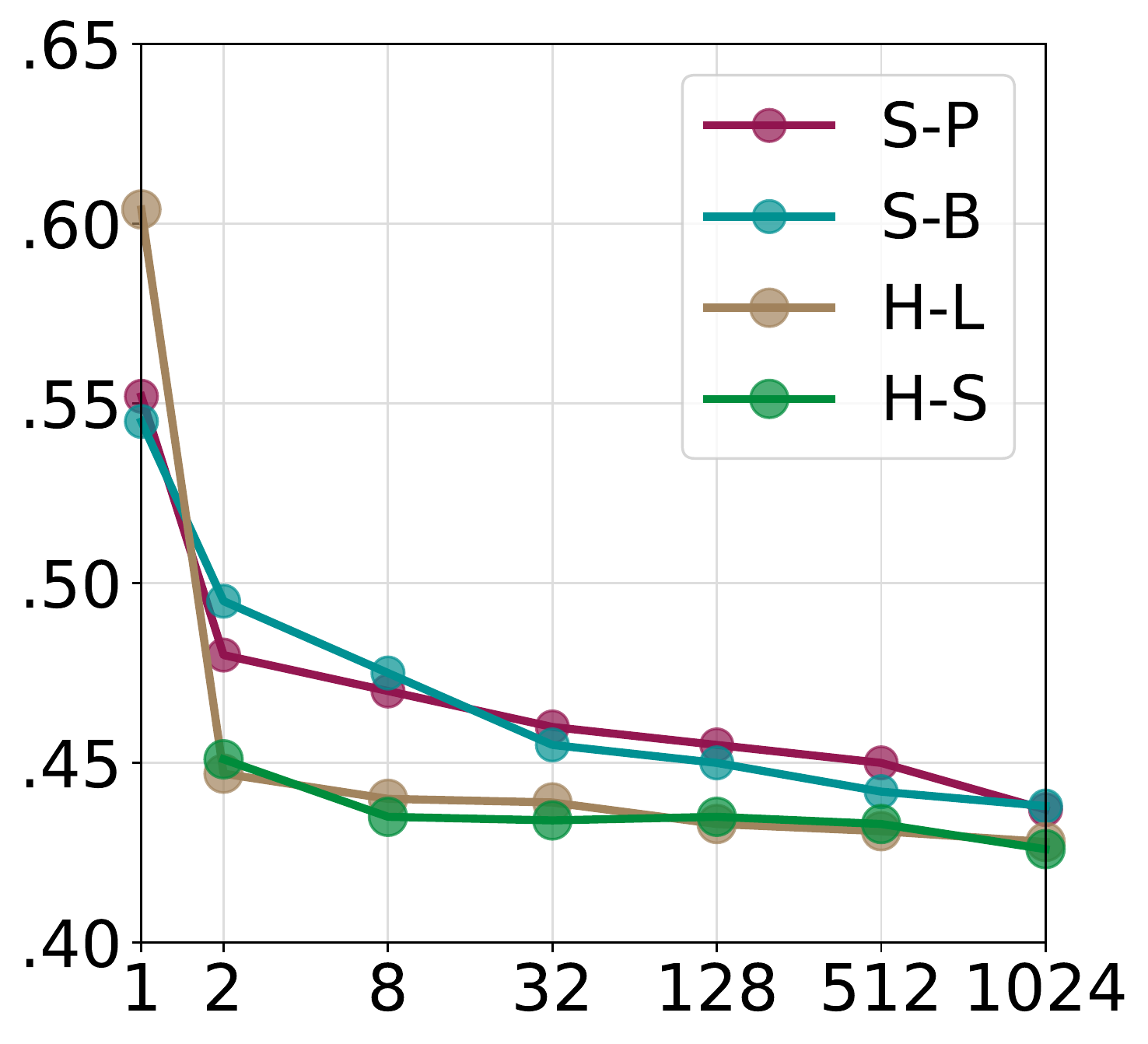}
    }
    \caption{Effects of the hyperparameters and the feature dimension on the performance of CPL. In each subfigure, the y-axis on the left represents the metric of MAE, and the y-axis on the right (if exists) represents the metric of accuracy.}
    \label{fig:param}
\end{figure*}

\subsection{Implementation Details}
\label{sec:implementation_details}

\vspace{0.3em}
\noindent\textbf{Feature Extractor.}
VGG-16 \cite{simonyan2014very} pretrained on ImageNet \cite{deng2009imagenet} is used as the feature extractor.
The last \textit{fc} layer of VGG-16 is replaced with a \textit{fc} layer to map the raw feature from the default 4096 dimensions to our specified dimensions.

\vspace{0.3em}
\noindent\textbf{Proxies Learner.}
The parameters of the proxies learner are initialized by Xavier \cite{glorot2010understanding} normal distribution, and are trained together with the feature extractor.

\vspace{0.3em}
\noindent\textbf{Training.}
We employ AdamW \cite{adamw} as the optimizer.
Learning rates of feature extractor and proxies learner are set as 0.001 and 0.01.
The batch size is set as 32.
All models are trained with PyTorch \cite{paszke2019pytorch} for 48 epochs.
During all training epochs, the model which achieves minimum MAE on the validation set is selected.

\vspace{0.3em}
\noindent\textbf{Image Setting.}
In training, the images are argumented by randomly cropping, resizing to $224\times 224$ and randomly horizontal flipping.
In testing, the images are processed by resizing to $256 \times 256$ and center cropping to $224 \times 224$.

\vspace{0.3em}
\noindent\textbf{Hyperparameter Setting.}
Dimension of features and proxies are 512 for fair comparison with baseline methods.
$\alpha$, $\tau_p$, $\tau_b$, and $s$ are set as 6, 0.11, 0.13, and 6, respectively.

\subsection{Comparison to Other Methods}
\label{sec:comparison}

\begin{figure*}[t]
    \centering
    \includegraphics[width=2.09\columnwidth]{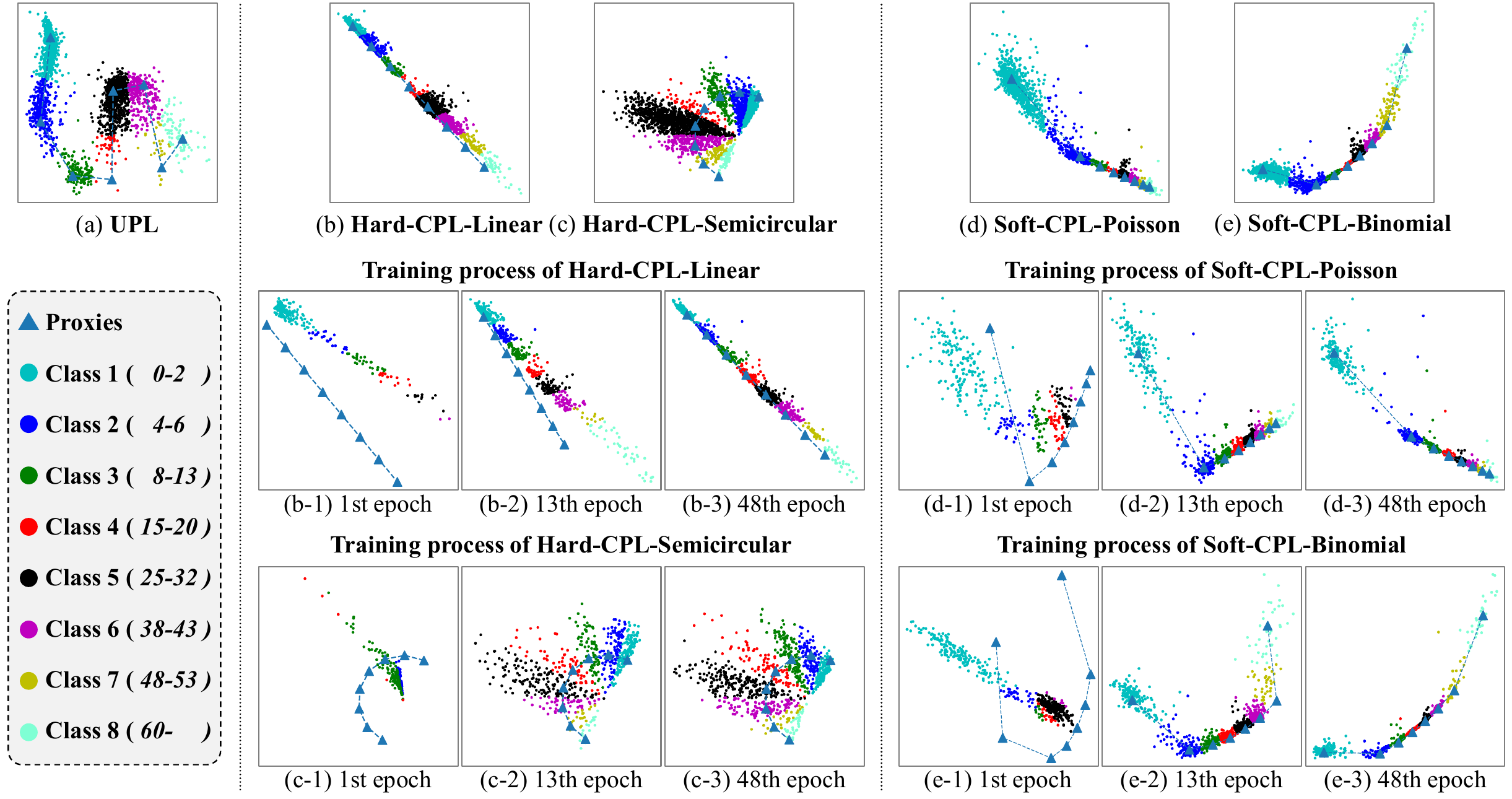}
    \caption{Visualization of CPL on the first fold of Adience Face dataset with eight age groups. 
    The dimension of features and proxies is set as 2.
    Only the features of correctly-predicted samples are plotted.
    The visualization on the test set are plotted on top.
    The visualization on validation set after 1st, 13th and 48th epochs are also plotted to illustrate the training processes.}
    \label{fig:visual}
\end{figure*}

Results on the three employed datasets are summarized in Table~\ref{tbl:comparison-historical-color-adience-face} and Table~\ref{tbl:comparison-image-aesthetics}.
In general, compared with all baseline methods, the proposed CPL achieves overall better performance.
Hard-CPL-Linear and Soft-CPL-Binomial with Euclidean distance achieve best MAE and best accuracy, respectively.
Among them, Hard-CPL-Linear outperforms the previous state-of-the-art (SOTA) method POEs \cite{li2021learning} on all three datasets by $1.03\%$, $1.1\%$, and $0.58\%$ higher on accuracy, $0.03$, $0.04$, and $0.007$ lower on MAE, respectively.
Soft-CPL-Binomial with Euclidean distance outperforms POEs by $3.28\%$, $1.6\%$, and $0.93\%$ higher on accuracy, $0.00$, $0.03$, and $0.001$ lower on MAE, respectively.
This demonstrates the effectiveness of our proposed CPL for the ordinal classification task.
Compared with UPL\footnote{UPL (Unconstrained Proxies Learning) encourages $P(\mathbf{f})$ to match the one-hot label distribution by the cross entropy loss.}, CPL achieves better results by a large margin, which means that explicitly controlling the class layout is effective for the ordinal classification task.

Among all settings of our CPL, Hard-CPL generally achieves better MAE than Soft-CPL, while Soft-CPL generally achieves better accuracy than Hard-CPL.
This may be because the hard layout constraints enforce the proxies to be placed in a more rigorous pre-defined ordinal layout, and thus can better catch the ordinal nature among classes.
For Soft-CPL, 
the strategy using Euclidean distance achieves better results than that using cosine similarity.
It means that Euclidean distance is more suitable for Soft-CPL.

\subsection{Model Analysis}

\vspace{0.3em}
\noindent\textbf{Effect of Scale Parameter $s$.}
To explore the effect of $s$ on the performance of H-S, we test the performance by varying the value of $s$ from $2$ to $10$ with step $2$ on the Adience Face dataset.
The results, which are summarized in Figure~\ref{fig:param}(a), show that both too small and too large $s$ deteriorate the performance.
The range of the logits in softmax function is $[-\infty,+\infty]$, while the range of cosine distance ($s=1$) is $[-1, 1]$.
Therefore, small $s$ makes the range of logits small, which reduces the discrimination ability between classes after softmax function.
Moreover, too large $s$ causes the target distribution to be close to the one-hot distribution after softmax, which further leads to worse performance because of less modeling of ordinal relationship.

\vspace{0.3em}
\noindent\textbf{Effect of Control Parameter $\tau_p$ and $\tau_b$.}
To explore the effects of $\tau_p$ and $\tau_b$ on the performance of Soft-CPL, we test the performance by varying the values of $\tau_p$ and $\tau_b$ from $0.07$ to $0.17$ with step $0.02$ on the Adience Face dataset.
The results, which are summarized in Figure~\ref{fig:param}(b), show that small $\tau$ leads to worse performance, while too large $\tau$ also deteriorates performance.
This is because smaller $\tau$ makes the target distribution tend to the one-hot vector, which lacks the modeling of ordinal relationship.
When $\tau$ becomes larger, the target distribution is more like uniform distribution, which reduces the discrimination ability between classes.

\vspace{0.3em}
\noindent\textbf{Effect of Tradeoff Parameter $\alpha$.} 
To explore the effect of tradeoff parameter $\alpha$, we test the performance by varying the value of $\alpha$ from $0$ to $12$ with step $2$ on Adience Face dataset. 
The results, which are summarized in Figure~\ref{fig:param}(c), show that smaller $\alpha$ leads to worse performance, while $\alpha$ larger than 6 achieves the stable performance.
This indicates that $\mathcal{L}_\mathrm{unimodal}$ can significantly help Soft-CPL to catch the ordinal nature.

\vspace{0.3em}
\noindent\textbf{Effect of Feature Dimension $d$.}
To explore the effect of feature dimension $d$, we test the performance by varying the value of $d$ from $1$ to $2048$ on Adience Face dataset.
The MAE results are summarized in Figure~\ref{fig:param}(d).
For Soft-CPL, MAE decreases with the increase of feature dimension.
Especially from $d=1$ to $d=2$, performance particularly improves.
It means Soft-CPL is more sensitive to feature dimension.
For Hard-CPL, MAE results are more stable under most feature dimensions, excluding $d=1$.

\subsection{Visualization}


\vspace{0.3em}
\noindent\textbf{Visualization of Hard-CPL.}
The visualization of Hard-CPL are summarized in Figure~\ref{fig:visual}(b) and Figure~\ref{fig:visual}(c).
Because of hard layout constraint, proxies and feature clusters are both arranged in expected ordinal layouts.
Each proxy is almost the centroid of corresponding feature cluster in H-L, while in H-S each proxy is roughly located in the central angle of the corresponding feature cluster.
The visualization of training processes are summarized from Figure~\ref{fig:visual}(d-1) to Figure~\ref{fig:visual}(e-3).
For H-L, Figure~\ref{fig:visual}(d-1) shows that clusters and proxies are getting closer until they completely coincide.
For H-S, Figure~\ref{fig:visual}(c-1) to Figure~\ref{fig:visual}(e-3) show that the angle between the feature and the target proxy decreases gradually.

\vspace{0.3em}
\noindent\textbf{Visualization of Soft-CPL.}
The visualization of Soft-CPL are summarized in Figure~\ref{fig:visual}(d) and Figure~\ref{fig:visual}(e).
For S-P and S-B, proxies and feature clusters are both arranged in expected ordinal layouts,
where different unimodal functions produce different relaxed ordinal layouts (i.e., with different shapes).
Each proxy is almost the centroid of the corresponding feature cluster.
The visualization of training processes are summarized from Figure~\ref{fig:visual}(b-1) to Figure~\ref{fig:visual}(c-3).
We can learn that the number of correctly classified features gradually increases, and the features of same-class samples are clustered more closely.

\section{Conclusion}

In this paper, we aim to learn a feature space specific to ordinal classification by explicitly constraining the layout of samples in feature space.
To this end, we propose the constrained proxies learning method.
From the perspectives of constraining the proxies layout in both hard way and soft way, we explore two strategies, \textit{i.e.}, Hard-CPL and Soft-CPL.
Hard-CPL directly controls the generation of proxies to force them to be placed in a strict linear layout or semicircular layout.
Soft-CPL constrains that the proxy layout should always produce unimodal proxy-to-proxies similarity distribution for each proxy.
We conduct experiments on three widely-used datasets of ordinal classification, and the experimental results demonstrate the effectiveness of the proposed CPL method.

\section{Acknowledgments}

This work is supported by National Nature Science Foundation of China under Grants Nos. 61972192, 62172208, 61906085, 41972111.
This work is partially supported by Collaborative Innovation Center of Novel Software Technology and Industrialization.

\bibliography{aaai23}

\appendix
\setcounter{secnumdepth}{1}

\section{Pseudo Code of CPL}

The pseudo code of our CPL can be seen in Algorithm~\ref{alg:sp}. 

\begin{algorithm}[ht]
    \caption{The Pseudo Code of CPL}\label{alg:sp}
    \begin{algorithmic}
        \STATE{\textbf{Input:} The training set $\mathcal{D}_\mathrm{Tr}$ and the validation set $\mathcal{D}_\mathrm{Val}$}
        \STATE{\textbf{Output:} The class of each sample in the test set $\mathcal{D}_\mathrm{Te}$}
        \STATE{\color{blue} $\vartriangleright$ \textbf{Training Stage}}
        \STATE{Initialize the model;}
        \FOR{each epoch}
        \FOR{each batch sampled from $\mathcal{D}_\mathrm{Tr}$}
        \STATE{Forward and calculate the loss (i.e., $\mathcal{L}_H$ or $\mathcal{L}_S$);}
        \STATE{Back propagate and update the model;}
        \ENDFOR
        \STATE{Evaluate the performance of the model on $\mathcal{D}_\mathrm{Val}$;}
        \ENDFOR
        \STATE{\color{blue} $\vartriangleright$ \textbf{Inference Stage}}
        \STATE{Load the model parameters which achieves the best performance on $\mathcal{D}_\mathrm{Val}$;}
        \FOR{each sample in $\mathcal{D}_\mathrm{Te}$}
        \STATE{Forward and predict the class of the sample;}
        \ENDFOR
        \RETURN{predicted classes of samples in $\mathcal{D}_\mathrm{Te}$}
    \end{algorithmic}
\end{algorithm}

\section{More Unimodal Distributions for Soft-CPL}

In our Soft-CPL, two unimodal distributions, i.e., Poisson distribution and Binomial distribution, are considered as the unimodal smoothing function.
While other unimodal distributions are also applicable, such as exponential function \cite{DBLP:journals/ijon/LiuFKDXLY20} or triangular distribution.

For exponential function, the ordinal smoothing function $E(\cdot, \cdot)$ is formulated as 
\begin{equation}
    E(k,k^*)=\frac{\exp(-|k-k^*|/\tau_e)}{\sum_{j=1}^{K}\exp(-|j-k^*|/\tau_e)}
\end{equation}
where $\tau_e$ is a hyperparameter to control the variance of the distribution, which is set as 30.

For triangular distribution, the ordinal smoothing function $E(\cdot, \cdot)$ is formulated as 
\begin{equation}
    f(k,k^*)=a-\frac{(a-b)\cdot|k-k^*|}{\max(k^*,K-k^*-1)}
\end{equation}
\begin{equation}
    E(k,k^*)=f(k,k^*)/\sum_{j=0}^{K-1}f(j,k^*)
\end{equation}
where $a$ and $b$ are the hyperparameters to control the maximum and minimum values of $f(k,k^*)$, which are set as 0.9 and 0.1, respectively.

As shown in Table~\ref{tab:unimodal-distribution}, the performances of the exponential function and the triangular distribution are similar with the performances of the Poisson distribution and the Binomial distribution.
In general, trying to find more suitable unimodal distributions for Soft-CPL is a valuable direction of our further research.

\begin{table}[ht] \setlength{\tabcolsep}{6.4pt} \footnotesize
    \centering
    \begin{tabular}{lcccccc}
    \toprule
    & \multicolumn{2}{c}{\textbf{HC}} & \multicolumn{2}{c}{\textbf{AF}} & \multicolumn{2}{c}{\textbf{IA}} \\ \cmidrule(lr){2-3} \cmidrule(lr){4-5} \cmidrule(lr){6-7}
    & \textbf{Acc.} & \textbf{MAE} & \textbf{Acc.} & \textbf{MAE} & \textbf{Acc.} & \textbf{MAE} \\ \midrule
    \textbf{Euc (P)} & 57.28 & 0.65 & 61.3 & 0.45 & 72.90 & 0.287 \\
    \textbf{Euc (B)} & 57.96 & 0.66 & 62.1 & 0.44 & 73.37 & 0.286 \\ 
    \textbf{Euc (E)} & 57.24 & 0.65 & 61.4 & 0.45 & 72.85 & 0.288 \\
    \textbf{Euc (T)} & 57.25 & 0.66 & 61.3 & 0.45 & 73.42 & 0.287 \\ \midrule
    \textbf{Cos (P)} & 56.99 & 0.65 & 61.1 & 0.46 & 72.82 & 0.286 \\
    \textbf{Cos (B)} & 57.66 & 0.65 & 61.9 & 0.44 & 73.21 & 0.287 \\
    \textbf{Cos (E)} & 57.01 & 0.66 & 60.9 & 0.45 & 73.20 & 0.288 \\
    \textbf{Cos (T)} & 56.97 & 0.67 & 60.7 & 0.45 & 73.19 & 0.288 \\
    \bottomrule
    \end{tabular}
    \caption{Performances of different unimodal distributions on Soft-CPL. 
    HC, AF, and IA denote the Historical Color dataset, the Adience Face dataset, and the Image Aesthetics dataset, respectively.
    Euc and Cos denote using Euclidean distance and using cosine similarity.
    P, B, E, and T denote Poisson, Binomial, exponential, and triangular, respectively.}
    \label{tab:unimodal-distribution}
\end{table}

\section{Effect of Negative Euclidean Distance}

An intuitive similarity function for Euclidean distance is negative Euclidean distance, which is formulated as
\begin{equation}
    \mathrm{sim}_\mathrm{E}(\mathbf{f},\mathbf{p}_k)=-\Vert\mathbf{f}-\mathbf{p}_k\Vert
\end{equation}
But in our CPL, the similarity function for Euclidean distance is formulated as:
\begin{equation}
    \mathrm{sim}_\mathrm{E}(\mathbf{f},\mathbf{p}_k)=-\log(1+\Vert\mathbf{f}-\mathbf{p}_k\Vert^2)
\end{equation}
which is widely-used in the Euclidean-based metric learning methods \cite{zhang-etal-2021-supporting, pmlr-v48-xieb16}.
It has the property that $(1+||\mathbf{f}-\mathbf{p}_k||^2)^{-1}$ approaches an inverse square law for large distances. 
This makes the probabilities almost invariant for the large distance.
It leads to smooth gradient variation, and more stable optimization \cite{van2008visualizing}. 

To compare the performance of negative Euclidean distance with that of our similarity function in CPL, we conduct experiments on the three datasets.
As shown in Table~\ref{tab:negative-euclidean}, the performances of negative Euclidean distance is a little worse than that of our similarity function.
It indicates that our similarity function is a better choice for the CPL methods when using Euclidean distance.

\begin{table}[ht] \setlength{\tabcolsep}{6.9pt} \footnotesize
    \centering
    
    \begin{tabular}{lcccccc}
    \toprule
    & \multicolumn{2}{c}{\textbf{HC}} & \multicolumn{2}{c}{\textbf{AF}} & \multicolumn{2}{c}{\textbf{IA}} \\ \cmidrule(lr){2-3} \cmidrule(lr){4-5} \cmidrule(lr){6-7}
    & \textbf{Acc.} & \textbf{MAE} & \textbf{Acc.} & \textbf{MAE} & \textbf{Acc.} & \textbf{MAE} \\ \midrule
    \textbf{H-L}        & 55.71 & 0.63 & 61.6 & 0.43 & 73.02 & 0.280 \\
    \textbf{H-L $\dag$} & 53.99 & 0.69 & 61.4 & 0.46 & 72.75 & 0.321 \\ \midrule
    \textbf{S-P}        & 57.28 & 0.65 & 61.3 & 0.45 & 72.90 & 0.287 \\
    \textbf{S-P $\dag$} & 56.44 & 0.73 & 61.1 & 0.47 & 72.13 & 0.318 \\ \midrule
    \textbf{S-B}        & 57.96 & 0.66 & 62.1 & 0.44 & 73.37 & 0.286 \\
    \textbf{S-B $\dag$} & 56.84 & 0.71 & 61.7 & 0.47 & 72.99 & 0.311 \\
    \bottomrule
    \end{tabular}
    \caption{Comparison the performance of negative Euclidean distance with that of our similarity function in CPL. $\dag$ denotes using negative Euclidean distance. HC, AF, and IA denote the Historical Color dataset, the Adience Face dataset, and the Image Aesthetics dataset, respectively.}
    \label{tab:negative-euclidean}
\end{table}

\section{Effect of the Value of $\Vert v_0\Vert$ for H-L}

In H-L, the vector $v_0$ is learnable.
We conduct the experiments to study the effect of $\Vert v_0\Vert$ on the performance of H-L by fixing the $\Vert v_0\Vert$ with different values. 
The results in Table~\ref{tab:v0} show that the performance degrades a lot if we fix $\Vert v_0\Vert$ to some default numbers rather than make it learnable.

\begin{table}[ht] \setlength{\tabcolsep}{5.0pt} \footnotesize
    \centering
    
    \begin{tabular}{lcccccc}
    \toprule
    & \multicolumn{2}{c}{\textbf{HC }} & \multicolumn{2}{c}{\textbf{AF}} & \multicolumn{2}{c}{\textbf{IA}} \\ \cmidrule(lr){2-3} \cmidrule(lr){4-5} \cmidrule(lr){6-7}
    & \textbf{Acc.} & \textbf{MAE} & \textbf{Acc.} & \textbf{MAE} & \textbf{Acc.} & \textbf{MAE} \\ \midrule
    \textbf{learnable $v_0$} 
    & 55.71 & 0.63 & 61.6 & 0.43 & 73.02 & 0.280 \\\midrule 
    \textbf{Fix $\Vert v_0 \Vert = 1$} 
    & 52.01 & 0.73 & 58.1 & 0.53 & 67.99 & 0.369 \\
    \textbf{Fix $\Vert v_0 \Vert = 3$} 
    & 52.12 & 0.71 & 58.4 & 0.52 & 68.02 & 0.341 \\ 
    \textbf{Fix $\Vert v_0 \Vert = 5$}
    & 52.11 & 0.71 & 58.5 & 0.52 & 67.98 & 0.335 \\
    \textbf{Fix $\Vert v_0 \Vert = 7$} 
    & 52.31 & 0.69 & 58.9 & 0.49 & 68.04 & 0.334 \\
    \bottomrule
    \end{tabular}
    \caption{Performance of H-L by fixing the $v_0$ with different values. HC, AF, and IA denote the Historical Color dataset, the Adience Face dataset, and the Image Aesthetics dataset, respectively.}
    \label{tab:v0}
\end{table}

\section{Proxies Learner of Hard-CPL-Semicircular}

Because the proxies are all located in the plane determined by $\mathbf{v}_0$ and $\mathbf{v}_1$, the generated proxies are the linear combination of $\mathbf{v}_0$ and $\mathbf{v}_1$, which means
\begin{equation}
    \mathbf{p}_k=
    a\cdot\frac{\mathbf{v}_0}{\Vert\mathbf{v}_0\Vert} +
    b\cdot\frac{\mathbf{v}_1}{\Vert\mathbf{v}_1\Vert}
\end{equation}
where $a,b\in\mathbb{R}$ are the combination weights. 
Based on the law of sines,
\begin{equation}
    \frac{\Vert\mathbf{p}_k\Vert}{\sin(\pi-\gamma)}=
    \frac{b}{\sin k\beta}=
    \frac{a}{\sin (\gamma-k\beta)}
\end{equation}
Because $\Vert\mathbf{p}_k\Vert=1$, then
\begin{equation}
    a = \frac{\sin(\gamma-k\beta)}{\sin\gamma},\ 
    b = \frac{\sin k\beta}{\sin\gamma}
\end{equation}
Finally, we get
\begin{equation}
    \mathbf{p}_k=\frac{\sin(\gamma-k\beta)}{\sin\gamma} \cdot
    \frac{\mathbf{v}_0}{\Vert\mathbf{v}_0\Vert} +
    \frac{\sin k\beta}{\sin\gamma} \cdot
    \frac{\mathbf{v}_1}{\Vert\mathbf{v}_1\Vert}
\end{equation}

\section{More Visualization}

\begin{figure*}[ht]
    \centering
    \includegraphics[width=2.1\columnwidth]{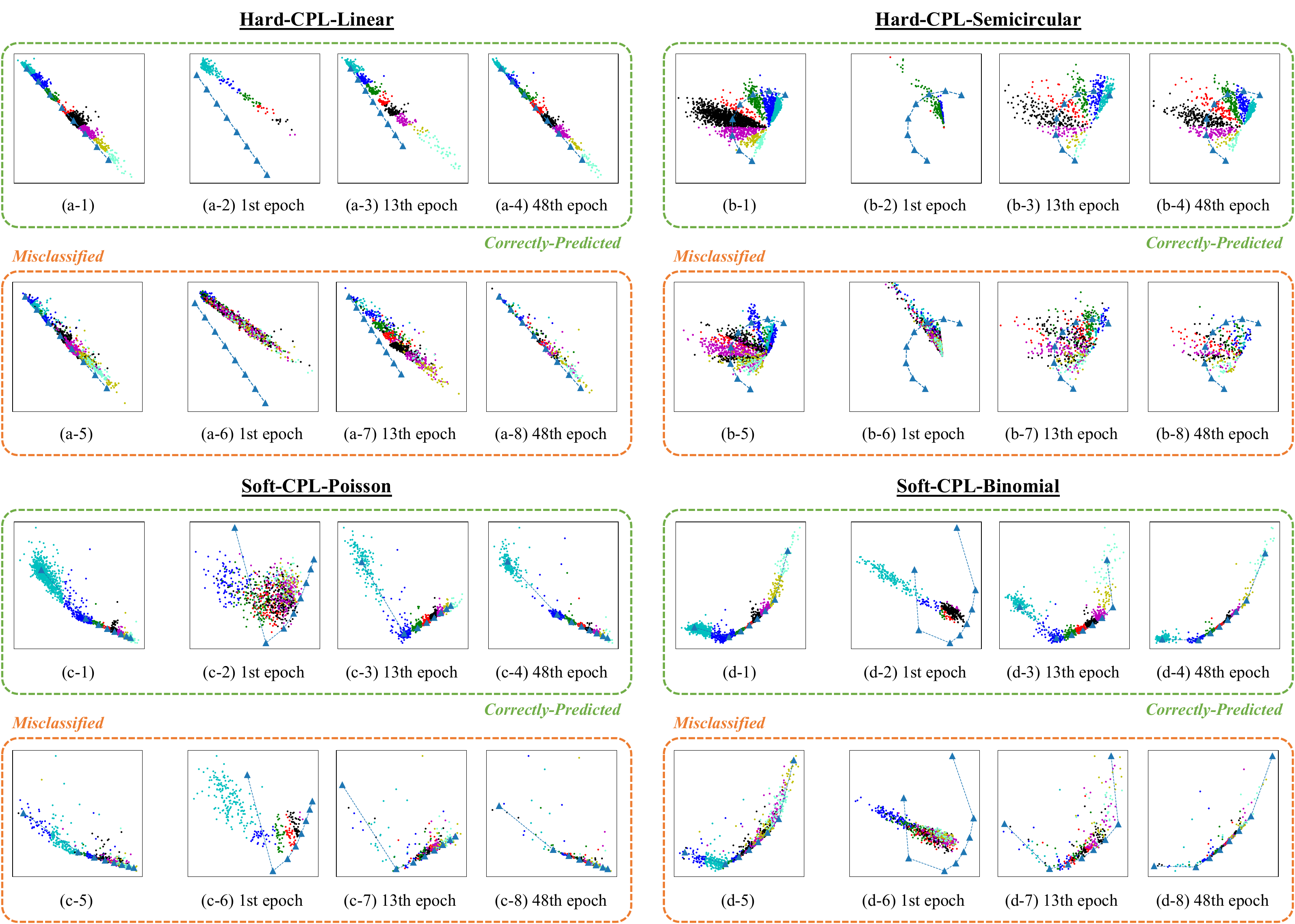}
    \caption{More Visualization of our proposed CPL on the first fold of Adience Face dataset with eight age groups.}
    \label{fig:more-visual}
\end{figure*}

In this section, we provide more visualization results in Figure~\ref{fig:more-visual}, including the features of correctly-predicted samples and misclassified samples. 
The features of misclassified samples are not evenly dispersed outside the decision boundary, but are mostly located in the decision areas of the adjacent ordinal classes. 
At the beginning of training, the features of misclassified samples are scattered disorderly in the space. 
However, with the progress of training, the features of misclassified samples are gradually reduced, and are mainly distributed in the decision areas of the adjacent ordinal classes.

\section{Trivial Collapsed Solution of H-L}

For the formulation of H-L, it seems obvious to contain a trivial solution which collapses all sample features and all proxies $\mathbf{p}_k$ into the zero point.
However, the collapse issue did not occur in the experiments. 
The reason is as follows.

In the KL divergence loss of H-L, the target distribution is provided by $Q(k^*)$ which will not be optimized, and $P(\mathbf{f})$ is the distribution to be optimized. 
Therefore, the condition the proxies collapse into the zero point is that $Q(k^*)$ is a uniform distribution. 
As long as $\Vert \mathbf{v}_0\Vert$ is not initialized as 0, $Q(k^*)$ will be the unimodal distribution, and proxies always do not collapse.

\end{document}